\documentclass[runningheads]{llncs}
\usepackage{graphicx}
\usepackage{amsmath,amssymb} 
\usepackage{color}
\usepackage{multirow}

\usepackage[width=122mm,left=12mm,paperwidth=146mm,height=193mm,top=12mm,paperheight=217mm]{geometry}

\begin{document}
\setlength{\abovecaptionskip}{5pt plus 3pt minus 2pt}

\pagestyle{headings}
\mainmatter

\title{Dense Image Representation with\\ Spatial Pyramid VLAD Coding of CNN\\ for Locally Robust Captioning} 

\titlerunning{Dense Image Representation with Spatial Pyramid VLAD}

\authorrunning{A. Shin M. Yamaguchi K. Ohnishi T. Harada}

\author{Andrew Shin \ Masataka Yamaguchi \ Katsunori Ohnishi \ Tatsuya Harada }
\institute{Graduate School of Information Science and Technology, The University of Tokyo}

\maketitle

\begin{abstract}
The workflow of extracting features from images using convolutional neural networks (CNN) and generating captions with recurrent neural networks (RNN) has become a de-facto standard for image captioning task. However, since CNN features are originally designed for classification task, it is mostly concerned with the main conspicuous element of the image, and often fails to correctly convey information on local, secondary elements. We propose to incorporate coding with vector of locally aggregated descriptors (VLAD) on spatial pyramid for CNN features of sub-regions in order to generate image representations that better reflect the local information of the images. Our results show that our method of compact VLAD coding can match CNN features with as little as 3\% of dimensionality and, when combined with spatial pyramid, it results in image captions that more accurately take local elements into account.
\keywords{CNN, VLAD, Spatial Pyramid, LSTM, Selective Search}
\end{abstract}

\section{Introduction}
Image captioning task has gained unprecedented attention with successful application of convolutional neural networks (CNN) and recurrent neural networks (RNN), especially long short-term memory (LSTM) units \cite{Karpathy,Vinyals,Xu,Donahue,Fang}. Such pipeline of extracting features from images using CNN, and mapping the representation to ground truth captions using RNN or LSTM has become a de-facto standard, employed by most recent works on image captioning task. With the current standard of CNN-LSTM pipeline, the novelty can come from either representation part (CNN), or learning and generation part (LSTM). We tackle the former part in this paper.

While CNN provides a powerful yet relatively compact representation of the image, it is noteworthy that CNNs are originally trained for classification of objects, with the goal of correctly identifying mostly a single, main object in the image. In image captioning task, however, it is frequently necessary to account not only for main objects in the image, but also for local, secondary objects. Although CNN mostly results in correct captioning with regards to the main object, it frequently comes up with incorrect captioning for local, secondary objects, as shown in Figure 1. This is somewhat natural in a sense that, as explained above, CNNs were originally trained for classification of main objects in the image.

In this paper, we introduce a novel application of spatial pyramid VLAD coding to CNN features at different sub-region levels, in order to generate more locally robust representation, and furthermore, more accurate captioning. VLAD has been popular coding method for compactly representing images from a large-scale dataset. However, its drawback of discarding spatial information has also been pointed out. In order to compensate for this drawback while preserving compact representation of VLAD, spatial pyramid VLAD has been suggested, and we employ it to CNN features.

In the conventional approach, CNN features are extracted from the image in its entirety without explicitly dealing with local objects. On the other hand, in our model, CNN features are extracted from a large number of bounding boxes from sub-regions proposed by selective search, which are mostly oriented towards local objects. This way, features are extracted not only from the entire image, but from each object or region whose importance is likely to be neglected in the conventional way. 

We then cluster the CNN features into a number of codewords, and perform VLAD coding using the codewords. Such coding results in very compact representation of images, as little as 3\% of the CNN features, and yet shows comparable performances. We then implement VLAD coding at different regions of different levels, thus implementing spatial pyramid VLAD so that the spatial information of the features extracted can be preserved. By doing so, we generate captions that more accurately and frequently account for local elements of the image that have been overlooked.

We optimize our method with various settings to investigate the influence of parameters and to find the best-performing combination. We also compare our method to previous works, as well as combining our method with conventional approach. Experimental results show that our method can more accurately and frequently account for local objects than the conventional approach, frequently providing more details than even human-written ground truth captions. 

Our main contributions comprise 1) showing that VLAD coding of CNN features from sub-regions can represent the images far more compactly than CNN features from the whole image, 2) combining it with spatial pyramid to account for spatial information, and 3) applying it to image captioning task to accomplish generation of more accurate and locally robust captions.

\begin{figure}[t!]
\vspace*{0mm}
\begin{center}
\includegraphics[width=9cm,height=2cm]{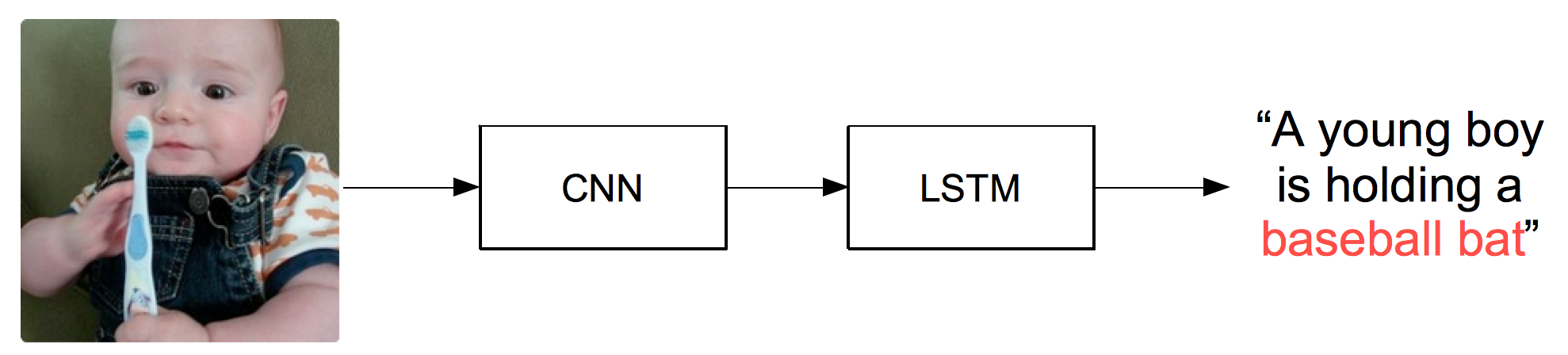}
\caption{Example of incorrectly captioned local objects using conventional approach}
\label{fig:boat1}
\end{center}
\vspace{-8mm}
\end{figure}

\section{Related Work}
A majority of recent work on image captioning task have been dominated by the usage of convolutional and recurrent neural networks for feature extraction and caption generation respectively, although with substantial variations. 

Inspired by statistical machine translation, Vinyals et al. \cite{Vinyals} built a model in which the encoder RNN for source sentences is replaced by CNN features of images. LSTM was employed as a generative RNN of non-linear function. This workflow of feature extraction using CNN followed by caption generation using LSTM builds a foundation upon which most recent image captioning works are based. 

Xu et al. \cite{Xu} took a similar workflow, but introduced attention-based model using standard back-propagation techniques, which learns to update the saliency while generating corresponding words. 

Donahue et al. \cite{Donahue} expanded the CNN-LSTM architecture to activity recognition and video recognition by building long-term recurrent convolutional networks (LRCNs). Time-varying inputs are processed by CNN whose outputs are fed to a stack of LSTMs.

Fang et al. \cite{Fang} took a more linguistically inspired approach by training visual detectors for words with multiple instance learning, which learns to extract nouns, verbs, and adjectives from regions in the image. Maximum-entropy language model generates a set of candidates, which are re-ranked by sentence-level features and deep multimodal similarity model. Our model differs from this work in that we explicitly take local objects into consideration rather than approximating from multiple instances.

\begin{figure*}[t!]
\centering
\begin{tabular}{cc}
{\includegraphics[width=4.2cm]{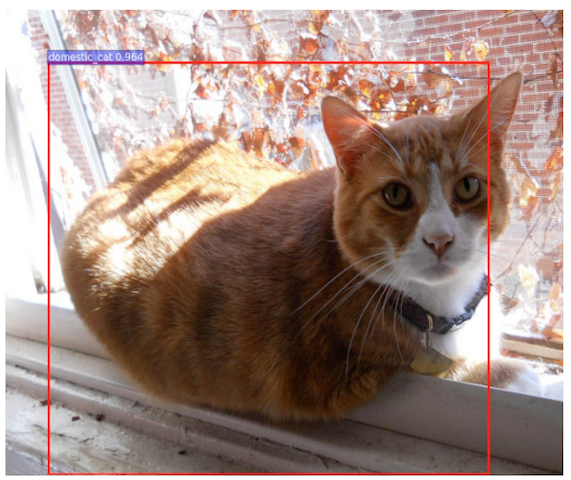}} \hspace{5.5cm}& 
{\includegraphics[width=4.2cm]{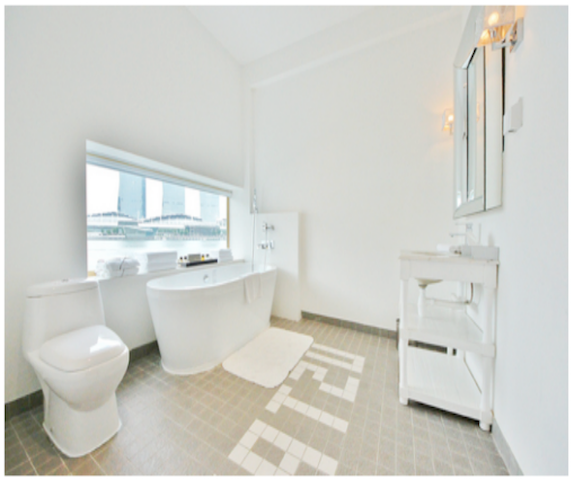}} \\
\end{tabular}
\caption{Examples of object detections using RCNN. Although it works well for objects present in 200 classes in the dataset (e.g.\ \textit{cat} in the image on the left-hand side), it frequently fails to detect objects that are not present in the dataset (e.g.\ \textit{toilet}, \textit{bathtub}, \textit{window} in the image on the right-hand side not detected). }
\label{fig:captions}
\vspace{-1ex}
\end{figure*}

Most of the above-mentioned works have represented images with CNN features extracted from the whole image, usually without paying explicit attention to local objects. In that regard, Johnson et al. \cite{DenseCap} show similar motivation with our work. They introduced DenseCap, which attempts to localize the regions in the image and incorporate it into captioning. However, their dense localization layer is trained on a dataset whose construction process is highly costly, with manual box-setting and labelling on crowdsourcing. Our method does not involve any manual labelling, and can work with any existing dataset.

Karpathy et al. \cite{Karpathy} exploited multimodal RNN to generate descriptions of image regions, aided by the alignment model of CNN over image regions and bidirectional RNN over sentences, which are intermingled via a multimodal embedding. This model also does not explicitly take local regions into representation, and instead relies on region convolutional neural network (RCNN) \cite{RCNN} to detect objects. However, since RCNN model is fine-tuned on only 200 classes of objects on PASCAL dataset \cite{Pascal}, it frequently fails to detect objects present in the image if those objects are not in the object classes of the dataset. Figure 2 shows examples of object detection using RCNN. On the other hand, since our model relies on selective search for object detection and region proposal, it is not limited by the number of object classes in the training dataset.

Vector of locally aggregated descriptors (VLAD) was first introduced by Jegou et al. \cite{VLAD}, as an efficient model to compactly represent images in a large-scale dataset, and has been a popular coding method for images, along with Fisher Vector \cite{FV}. They used the simple L2 normalization method for normalizing VLAD descriptors. Arandjelovic et al. \cite{AllVLAD} demonstrated that intra-normalization and recording multiple VLADs for an image, along with vocabulary adaptation, can further enhance the performance of VLAD.

As a method to approximate global non-invariant geometric statistics, Lazebnik et al. \cite{SPM} introduced spatial pyramid matching technique, a simple extension of bag-of-features representation, in which histograms for local features are aggregated in each sub-region. 

Although spatial pyramid can find useful global features from each level, it has been reported to be weak at high geometric variability, necessitating a combination with invariant features. On the other hand, VLAD coding is usually performed on locally invariable descriptors, such as SIFT, yet it does not preserve spatial information. In order to compensate for these mutual weaknesses, Zhou et al. \cite{SPVLAD} introduced spatial pyramid VLAD, which combines the two. This technique will play a central role in the method introduced in our paper.

Sanchez et al. \cite{augment} showed a far simpler approach for taking spatial information into account, by simply incorporating the coordinate information into the feature vector and augmenting it. We will also examine this approach and compare it to our model in Section 4.

Some previous works \cite{ECCV,CVPR,arXiv} have applied similar methods to ours by extracting deep activation features from local patches at multiple scales, and coding them with VLAD or Fisher Vector. However, previous works mainly dealt with scene classification, texture recognition, and object classification, in which the necessity for explicitly dealing with local objects and spatial information is less pronounced. On the contrary, image captioning task requires that local objects be very clearly reflected in the captions, which we manage by novel application of spatial pyramid matching to region-level CNN features. To the best of our knowledge, our work is the first to apply such workflow to image captioning task.

\section{Model}
In this section, we describe each stage of our workflow, starting with region-based feature extraction using selective search, and ending with caption generation using LSTM. Pictorial description of the overall workflow of our model is shown in Figure 3.

\subsection{Region-Based Feature Extraction}
We first obtain a set of region proposals from images, using selective search \cite{SS}. Selective search starts by superpixel segmentation, and then proceeds with hierarchical grouping of regions in a bottom-up manner, in which neighbouring regions are combined iteratively. We used the ``fast mode'' of selective search, in which HSV and Lab colorspaces are employed, and other measures, such as size of region and similarity between neighbouring regions, are also taken into consideration. As was discussed in Section 2, the largest benefit of using selected search is that it is not limited by the number of object classes, which was the case for object detection using RCNN and other object detection methods based on datasets.

We then extract CNN features from \textit{all} regions proposed by selective search. The rationale behind extracting CNN features from region proposals is that, since regions now tightly encompass particular objects, CNN features from the regions will be highly representative of that particular object. The motivation for feature extraction from \textit{all} regions instead of running non-maximum suppression and reducing the number of regions consists mainly of two reasons. First, as we will see later, CNN features will go through spatial coding, in which case an insufficient amount of region samples can cause data sparsity problem. Second, it is intuitive that conspicuous objects will have multiple proposals of different sizes, so that the influence of such objects will remain strong even after coding, and are likely to be reflected in captions.

Until recently, extracting CNN features from all region proposals would have been an impractical idea due to time issue. However, recent releases of high-speed, region-based variation of CNN, such as Fast-RCNN \cite{FastRCNN}, have made the idea feasible. Although Fast-RCNN was originally designed for detection task, we took advantage of it to perform high-speed feature extraction. Instead of provided network models trained for detection task, we used the model trained for classification on ImageNet \cite{ImageNet}, specifically VGG network \cite{VGG}, and extracted 4096-dimensional features from the second fully-connected (fc7) layer.

\begin{figure}[t!]
\begin{center}
\includegraphics[width=12cm,height=7cm]{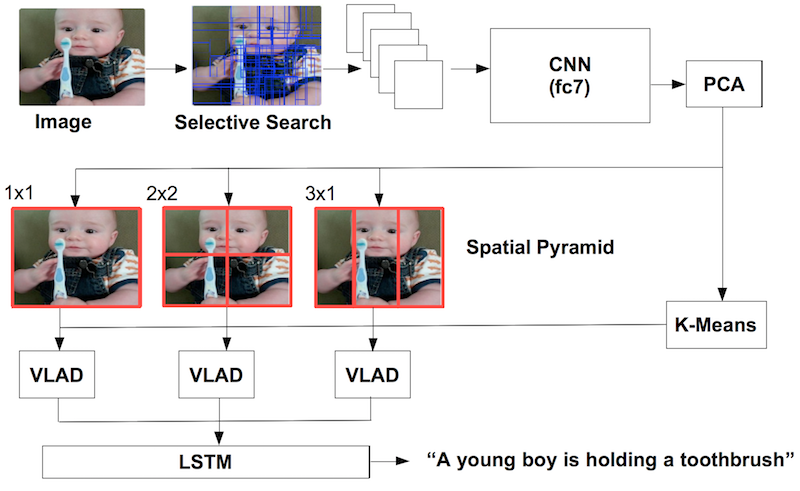}
\caption{Overall workflow of our model.}
\label{fig:boat1}
\end{center}
\vspace{-5mm}
\end{figure}

\subsection{VLAD Coding}
Since we will eventually have to perform VLAD coding of the features separately on each grid of spatial pyramid, 4096-dimensions will be too large and even redundant. We thus performed dimensionality reduction with PCA on CNN features extracted from the regions. We trained the PCA with features of 250k randomly sampled regions, and separately performed reduction to 128, 256, 512, and 1024-dimensions.

We then performed codeword learning with K-means. The number of clusters included 1, 2, 4, 8, 64. Centroids were initialized with K-means++ \cite{KMPP}, as random Gaussian initialization resulted in skewed clustering. 

Based on the codewords learned using K-means, we perform the VLAD coding on CNN features obtained after PCA, with signed square rooting normalization as in \cite{Square}. Dimensionality of the resulting vector at this point will be the dimensionality of CNN features obtained after PCA times the number of clusters.

\subsection{Spatial Pyramid}
Although VLAD encoding is known to perform well on preserving locally invariant features, it is at the cost of discarding spatial information. Previous works have thus proposed Spatial Pyramid VLAD, in which VLAD coding is performed at multiple levels of different sizes, going from coarse to fine sub-regions, inspired by spatial pyramid matching.

In order to determine which grid a region belongs to, we simply locate the center of the region. An alternative would be to assign a region to all overlapping grids. However, our examination of the alternative resulted in grids not being much different from each other, and thus not discriminative enough. This may be attributed to the fact that many conspicuous elements in the images are frequently large in size, thus occupying multiple grids, and consequently making the grids all similar. We thus resorted to the center of the region for its grid assignment, as it is also more concurrent with our initial motivation of preserving local information.

Since previous works \cite{SPM} have reported that levels beyond three obtain only an insignificant amount of improvement at the cost of enlarged dimensions, we also set our spatial pyramid at three levels; $1\times1$, $2\times2$, and $3\times1$ (left, middle, right). Thus, using up to second level of the pyramid will result in representations of the image at 5 different sizes or locations, and up to third level will have representations at 8 different sizes or locations per image.

\subsection{Caption Generation}
Since the main motivation of this paper is to tackle the representation part of the image captioning process, we generally follow the conventional approach for caption generation part, applying LSTM to our representation of the images and ground truth annotations. So-called ``vanilla" architecture for LSTM was used \cite{KarpathyLSTM}. 
Word vectors were trained with random initialization, and sigmoid function is used for non-linearity throughout all gates except along with hyperbolic tangent for memory cell update. The number of iterations was fixed to 35,000 in all experiments, and beam size of 1 was used.

\section{Experiment}

\subsection{Setting}
We apply our method to de-facto standard dataset for image captioning task, namely MS COCO \cite{MSCOCO}. Train, validation, and test split add up to roughly 160,000 images, and running selective search on the entire dataset resulted in 61.3M region proposals, roughly 385 regions per image, including the entire image itself. Thus, each cell in spatial pyramid, except for $1\times1$, contains 96 to 128 sub-regions on average. Starting from these region proposals, we follow the procedures described in Section 3 to generate captions.

We compare the performance of our proposed model with that of baseline, in which 4096-dimensional CNN features are extracted from the entire image, and are then inserted to LSTM as input with no further preprocessing.

\subsection{Parameter Validation}

One issue with spatial pyramid VLAD coding with CNN is that there are a number of factors to consider that can potentially contribute to a large increase in dimensionality of the final representation; namely, dimensionality of reduced CNN features, number of clusters (codewords), and level of pyramid. Even if performance increases, excessively high dimensionality would be impractical. We would thus like to take into consideration an appropriate tradeoff between performance and dimensionality throughout our experiment, and performed a number of validations to set up appropriate parameters.

We first examined the influence of dimensionality of the CNN features after PCA, \textit{ceteris paribus}. The number of codewords was fixed to 1, and spatial pyramid was not employed (or, to put it in a different way, only $1\times1$ grid was applied). We varied the dimensionality of CNN features as 128, 256, 512, and 1024, to which VLAD coding was applied with one cluster. The original dimensionality of CNN features prior to PCA, which is 4096, was not employed since it will cause the the final representation to be impractically large, when combined with multiple clusters and spatial pyramid.

Secondly, we examined the influence of the number of clusters, \textit{ceteris paribus}. Dimensionality of the CNN features was fixed to 256, which achieved the best performance in the first validation, and spatial pyramid was not employed. We varied the number of clusters as 1, 2, 4, 8, and 64, with which VLAD coding was performed upon 256-dimensional CNN features.

\begin{table*}[t!]
\small
\begin{center}
\caption{Performances on BLEU with varying dimensionalities of CNN features after PCA (1 cluster, no spatial pyramid)}
\begin{tabular}{@{ \ }c@{ \ \ }|@{ \ \ }c@{ \ \ }c@{ \ \ }c@{ \ \ }c@{ \ \ }c@{ \ }}
\hline \bf Dimension & \bf BLEU-1 & \bf BLEU-2 & \bf BLEU-3  &\bf BLEU-4 \\ \hline
128 & 59.3 & 39.3 & 25.4 & 16.9 \\ \hline 
256 & 59.3 & 39.4 & 25.5 & 17.0 \\ \hline
512 & 51.2 & 30.8 & 17.9 &  10.9 \\ \hline
1024 & 48.0 & 25.4 & 13.2 & 7.8  \\ \hline
\end{tabular}
\label{table:human}
\end{center}
\vspace{-1ex}
\end{table*}

\begin{table*}[t!]
\small
\begin{center}
\caption{Performances on BLEU with varying number of clusters (256 dimension, no spatial pyramid)}
\begin{tabular}{@{ \ }c@{ \ \ }|@{ \ \ }c@{ \ \ }c@{ \ \ }c@{ \ \ }c@{ \ \ }c@{ \ }}
\hline \bf Clusters & \bf BLEU-1 & \bf BLEU-2 & \bf BLEU-3  &\bf BLEU-4 \\ \hline
1 & 59.3 & 39.4 & 25.5 & 16.9 \\ \hline
2 & 58.7 & 38.8 & 24.9 & 16.3 \\ \hline
4 & 60.7 & 41.0 &27.3 & 18.4  \\ \hline  
8 & 57.7 & 36.4 & 22.5 &  14.6 \\ \hline
64 & 46.4 & 25.0 & 10.4 &4.4 \\ \hline
\end{tabular}
\label{table:human}
\end{center}
\vspace{-4ex}
\end{table*}

Table 1 shows the performances of our model with various dimensionality of CNN features after PCA, using BLEU \cite{BLEU} as evaluation metric. Notably, lower dimensionality outperforms higher ones by a considerable margin. This may initially come as a surprise, but it makes sense with scrupulous inspection of our pipeline. Note that CNN features in our method are extracted from small regions suggested by selective search, which mostly contain objects at a large scale, as opposed to the ``whole'' images containing various objects and components at varying scales. Thus, much fewer dimensionality is needed to correctly classify and represent the objects, as compact as 128. Furthermore, since clusters are obtained from these features, less compact representation with high dimensionality is likely to result in noisy clusters, consequently leading to noisy VLAD coding, which negatively affects the accuracy of captions.

Surprisingly, 256-dimensional coded representation, even with only one cluster and no spatial pyramid, resulted in best performance, almost equal to the performance of 4096-dimensional CNN features used in conventional image captioning task. Further-reducing the dimensionality to 128 resulted in very slight decrease, but still comparable to 256-dimension features and CNN features, despite being only 1/32 of its size. This demonstrates that VLAD coding of CNN features from region proposals contains highly discriminative ability, while being very compact.

Table 2 shows the performances of our model with various numbers of clusters. The differences in performance between lower number of clusters are relatively small, while larger number with 64 clusters noticeably degrades the performance. Similarly to the case of dimensionality, a large number of clusters results in sparse clustering, where many clusters end up with no vector assigned to it. In fact, inspecting the resulting VLAD-coded vectors with 64 clusters shows that it contains a large number of zeros, which must have become noise and hindered the learning. Increasing the number of iterations is likely to improve the performance to similar levels as lower numbers of clusters, but it indicates that its convergence is much slower.

Although using only one or two clusters resulted in high performances as well, it would not be much different from average pooling, and would thus not utilize the benefit of VLAD coding to full extent. Thus, in the following experiments for comparison to previous works, we mostly proceed with the combination of 256-dimensional CNN features, 4 or 8 clusters, and spatial pyramid of level 2 or 3, although the results from a few other combinations will also be reported for reference.

\subsection{Additional Setup 1: Feature Augmentation}

In object detection and classification literature, some alternatives to spatial pyramids have been proposed. We examine one of such proposals in order to examine whether, and to what extent, local information can be preserved without using spatial pyramids. 

Sanchez et al. \cite{augment} have proposed a simple feature augmentation method, where coordinate information is concatenated directly to the vector of descriptor. Specifically, given 2D-coordinates of a region patch $m_t = {[m_{x,t},m_{y,t}]}^T$ with a descriptor $x_t$ of size $D$, which in our case corresponds to CNN features, and the patch scale $\sigma_t$, where the image is of size H and W, we augment the dimension of the descriptor by 3, resulting in a new vector $\hat{x}_t\in\mathbb{R}^{D+3}$ as follows:

\begin{equation}
  \hat{x}_t = 
  \begin{pmatrix}
    x_t \\
    m_{x,t}/W - 0.5 \\
    m_{y,t}/H - 0.5 \\
    \log\sigma_t - \log\sqrt{WH}
  \end{pmatrix}
\end{equation}

Thus, it accounts for location information of each region implicitly in the feature vector, rather than explicitly dividing regions and generating separate representations. The obvious benefit of this approach is the simplicity of its implementation. 

As an additional experiment, we implement this method to CNN features with reduced dimensionality, and compare its performance to our proposed model. The best-performing set of parameters as in our model was employed; 256-dimension CNN features with 4 clusters, but no spatial pyramid. Thus, features are augmented to 259-dimension, and the resulting final representation becomes $259\times4$=1036 dimensions, roughly a quarter of the dimensionality of conventional CNN features.

\subsection{Additional Setup 2: Ours + CNN (Whole)}

We also examine a combination of our method with the conventional approach, in which CNN features are extracted from the image in its entirety. CNN features extracted in the conventional approach are holistic features, so to speak, whereas our model is more concerned with specific objects in sub-regions, taking their location into consideration. Thus, the motivation is to take into consideration both generals and particulars of the images. 

For CNN features in conventional approach, we extracted activations from second fully-connected (fc7) layer of 4096-dimension with VGG-16 network trained on ImageNet using Caffe framework \cite{Caffe}. For our model, 256-dimensional CNN features with 4 clusters and spatial pyramid up to level 2 were employed. Combining the two adds up to 9216 dimensional vector per image.

\subsection{Evaluation}

Table 3 and Table 4 summarize the results from our model and previous works, with various combinations of parameters. We tested the performance on a number of widely used evaluation metrics, including BLEU, METEOR \cite {METEOR}, Cider \cite{Cider}, and perplexity. Apart from perplexity, which is calculated from the inverse probability of the words, all metrics give higher score for better results. Note that the scores for CNN features following the conventional approach are from our own experiment under the same condition for fair comparison, and other papers have reported varying, sometimes higher results with variations in their methods, although these varying scores are mostly in a close range. Although enhancing the performance of LSTM is out of scope of this paper, it is an active research area, and replacing our LSTM with more advanced versions is very likely to boost the overall performance of all models.

Most of the variations attempted achieve performances very close to CNN features, especially with the best-performing combination outperforming CNN features at BLEU-4. Note that BLEU-4 is computed from higher-order of n-grams than others, and is thus frequently employed as the primary source of evaluation metric \cite{Vinyals,Workshop}, since it is considered to better-indicate the overall semantic similarity. It thus shows that accounting for local objects as in our model enhances the overall semantic accuracy.

\begin{table*}[t!]
\small
\begin{center}
\caption{Performances of each model on BLEU.}
\begin{tabular}{c|c|c|c|cccccc}
\hline \bf CNN Dim. & \bf Cluster & \bf SP Lev. & \bf Total Dim.  & \bf BLEU-1 & \bf BLEU-2 & \bf BLEU-3& \bf BLEU-4 \\ \hline 
CNN (whole) & N/A & N/A & 4096 & 62.0 & 42.4 & 28.0 & 18.7  \\ \hline \hline
\multirow{6}{*}{256} & \multirow{3}{*}4 & 1 & 1024 & 60.7 & 41.0 &27.3 & 18.4  \\ 
&  & 2 & 5120 & 61.3 & 41.4 & 27.5 & \bf 18.9  \\ 
&  & 3 & 8192 & 61.1 & 40.8 & 26.9 & \bf 18.8\\ 
& \multirow{3}{*}8 & 1 & 2048 & 57.7 & 36.4 & 22.5 &  14.6 \\	
&  & 2 & 10240 & 58.5 & 37.4 & 23.7 &  15.5 \\ 
& & 3 & 16384 & 58.9 & 38.8 & 24.9 & 16.5 \\ \hline 
259 \cite{augment} & 4 & N/A & 1036 & 56.5 & 35.9 & 21.9 &  13.9  \\ \hline 
Ours+CNN & N/A & N/A & 9216 & 60.5 & 41.0 & 27.1 &  18.4  \\ \hline 
\end{tabular}
\label{table:scores}
\end{center}
\vspace{-4ex}
\end{table*}

 \begin{table*}[t!]
\small
\begin{center}
\caption{Performances of each model on Various Evaluation Metrics}
\begin{tabular}{c|c|c|c|cccccc}
\hline \bf CNN Dim. & \bf Cluster & \bf SP Lev. & \bf Total Dim.  & \bf METEOR & \bf Cider & \bf Perlexity \\ \hline 
CNN (whole) & N/A & N/A & 4096 & 12.1 & 62.3  &  13.45  \\ \hline \hline
\multirow{6}{*}{256} & \multirow{3}{*}4 & 1 & 1024 & 11.7 & 56.6  & 14.02  \\ 
&  & 2 & 5120 & 11.9 & 61.1  &  13.69 \\ 
&  & 3 & 8192 & 11.7 & 59.7  &  13.62\\ 
& \multirow{3}{*}8 & 1 & 2048 & 9.6 & 41.5  &  14.28 \\	
&  & 2 & 10240 & 10.0 & 45.1  &  13.84 \\ 
& & 3 & 16384 & 10.7 & 47.3  &  13.73 \\ \hline 
259 \cite{augment} & 4 & N/A & 1036 & 9.5 & 40.1  &  14.30  \\ \hline 
Ours+CNN & N/A & N/A & 9216 & 11.1 & 51.2  &  13.86  \\ \hline 
\end{tabular}
\label{table:scores}
\end{center}
\vspace{-3ex}
\end{table*}

Other evaluation metrics generally agree with BLEU. Overall, 256-dimensional CNN features with 4 clusters up to level 2 and 3 resulted in best performance. In all cases, models with spatial pyramids outperform those without, \textit{ceteris paribus}, which demonstrates that paying attention to local elements by dividing the images into sub-regions is able to reflect more detailed aspects of the images. Patterns observed in parameter validation in Section 4.2 mostly hold true, with compact dimensionality and not-too-large number of clusters performing better. 

Feature augmentation achieved reasonably high performance, but fell short of our model. If there are multiple regions covering the same objects, those regions have close feature vectors as well as close coordinates. Thus, when they are assigned to clusters, the explicitness of coordinate information is likely to become subdued to an insignificant extent. This again shows the importance of explicitly accounting for local objects in image representation.

In order to examine how different our captions are from original CNN features, we calculated BLEU score of our model with captions generated from CNN features as references, which resulted in 59.4/47.9/40.2/35.3. The score shows that, while both captions feature similar contents, they have considerable differences in their wordings and in their dealing of details.
\subsection{Discussion}
Figure 4 shows examples of images and captions generated by our model, CNN features, and the combination of two, along with ground truth. See Appendix for more examples.

\begin{figure*}
\centering
\begin{tabular}{cc}
{\includegraphics[width=4.2cm]{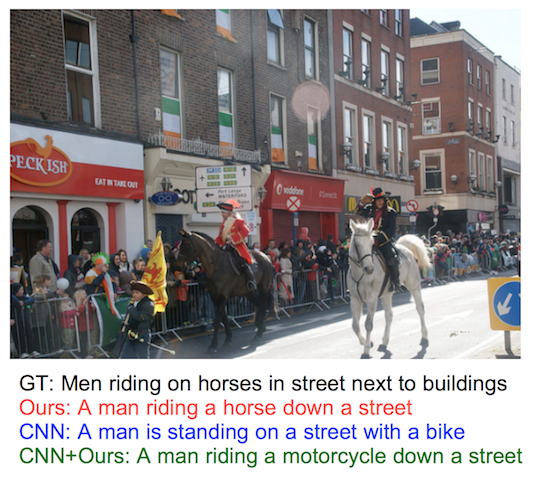}} \hspace{1.2cm} & 
{\includegraphics[width=4.7cm]{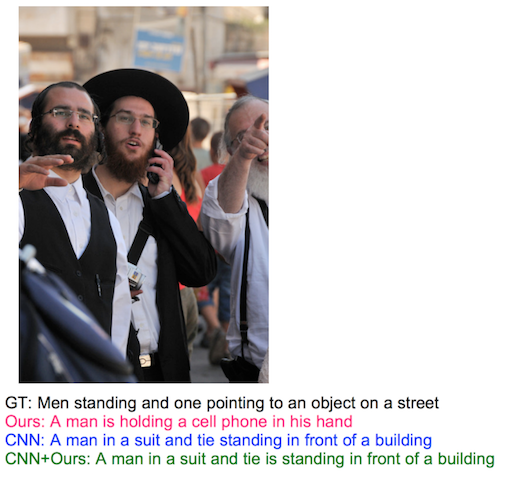}} \\ 
{\includegraphics[width=4.7cm]{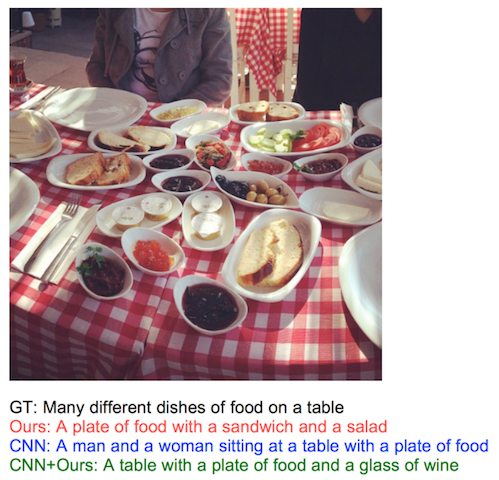}} \hspace{1.2cm}&
{\includegraphics[width=4.7cm]{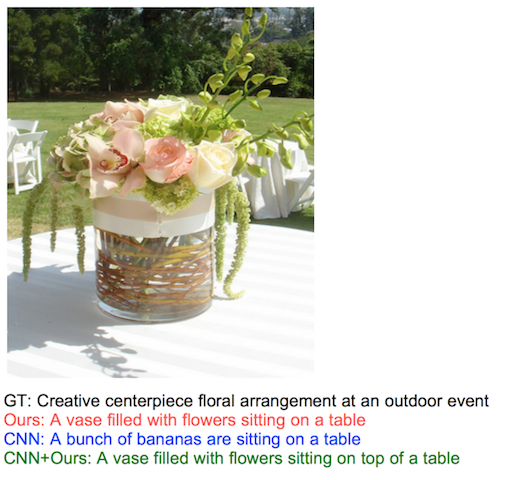}} \\
{\includegraphics[width=4.7cm]{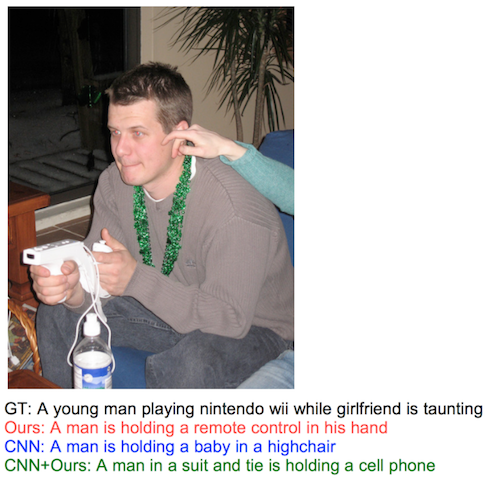}}\hspace{1.2cm}&
{\includegraphics[width=4.7cm]{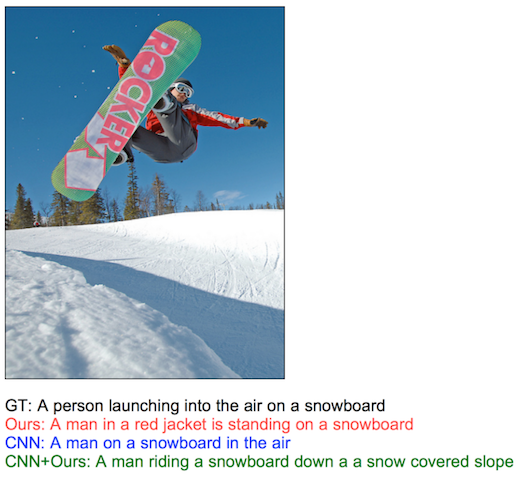}} \\
{\includegraphics[width=4.7cm]{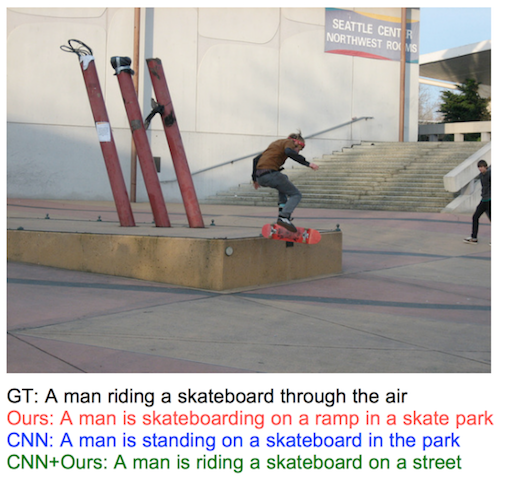}} \hspace{1.2cm}&
{\includegraphics[width=4.7cm]{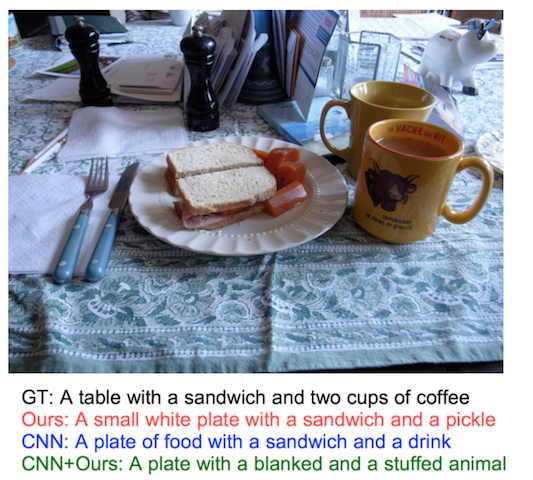}}\\
\end{tabular}
\caption{Examples of captions generated by each model, along with ground truth captions. Captions generated by our model frequently contain description of local, secondary objects, sometimes with even more details than ground truths.}
\label{fig:captions}
\vspace{-3ex}
\end{figure*}

Resulting captions show that our model does capture local, secondary objects more correctly and frequently than CNN features, often providing more details than even the human-written ground truth captions. It may potentially have put us at a disadvantage in terms of performance on similarity-based evaluation metrics, but it verifies that our model has successfully learned to apply mapping between local objects and their linguistic correspondences to new images, and that our motivation of capturing spatial information with spatial pyramid has succeeded to a plausible extent.

There were indeed cases where our models performed more poorly than CNN features. Such cases were mostly the ones in which it was hard to find any component other than main object in the image. Our model often talks about non-existent secondary object, or, in worse case, incorrectly describes the main object. As much as it can deal better with particular local objects when they are present, it turns out to be less efficient when there are no secondary objects so that segmenting the image into spatial pyramid becomes unnecessary. Figure 5 shows examples of such  cases.

Since our model better-deals with local objects, while original CNN features can handle main objects well, it seems intuitive to combine the two and expect balanced results. However, their performances on evaluation metrics were slightly lower than respective models, largely due to their large dimensionality, which requires more iterations to fully converge. The resulting captions seem to display somewhat mixed characteristics, slightly leaning more towards captions from CNN features in terms of contents. 

Since applying PCA to our VLAD-coded CNN features from sub-regions not only reduced the dimensionality but also enhanced the performance, one possible alternative would be to apply dimensionality reduction to original CNN features as well, and see whether it retains its discriminative strength. If successful, it will make the combination of two models more compact and thus more practical.

\begin{figure*}[t!]
\centering
\begin{tabular}{cc}
{\includegraphics[width=4.7cm]{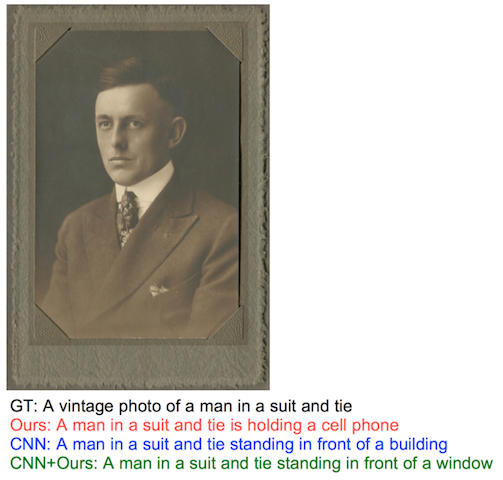}} \hspace{5.5cm}& 
{\includegraphics[width=4.7cm]{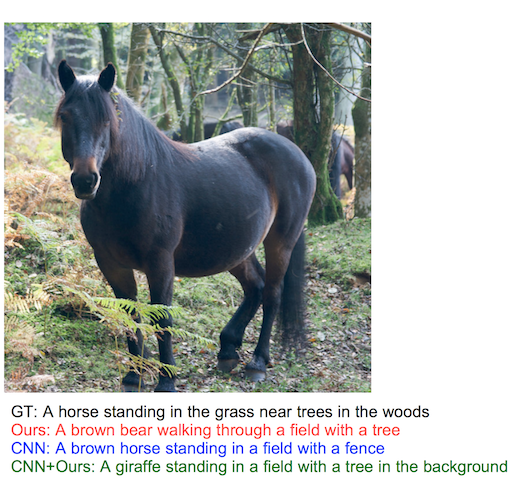}} \\
\end{tabular}
\caption{Examples in which captions generated by our model performed poorly. Many of them hardly included any secondary object and could not benefit much from segmentation by spatial pyramid.}
\label{fig:captions}
\vspace{-3ex}
\end{figure*}

\section{Conclusion}
We introduced a novel method  for image representation incorporating spatial pyramid VLAD to CNN features of sub-regions suggested by selective search, in order to generate more locally robust image captions. Our VLAD coding of CNN features, both with and without spatial pyramid, was able to achieve performance nearly equivalent to CNN features, while having much lower dimensionality, as little as 3\% of its size. We optimized our model via parameter validations, and learned that combination of low dimensionality after PCA with appropriate number of  clusters yields the best results. 

Combining spatial pyramid turned out to enhance performance not only on evaluation metrics, but in resulting captions dealing well with local objects. Our model more accurately and frequently accounts for local objects than previous methods, such as feature augmentation or conventional CNN representation for the whole image. It frequently dealt with local objects with more details than even the human-written ground truth captions. Our model did show weaknesses when there are no local elements so that spatial pyramid is hardly necessary, but a more balanced combination with conventional CNN features is likely to complement mutual weaknesses of respective models.

Since spatial pyramid has pre-determined division of cells that may not always correspond to the ideal localization of objects in the image, it may be helpful to build a spatial pyramid in which the size and location of the cells are determined by the results of region proposals followed by non-maximum suppression. This topic will be our immediate future work. 

Also, since we exclusively dealt with the representation part of the image captioning task, a novel approach to tackle the generation part of it would naturally be of interest.

\clearpage

\bibliographystyle{splncs}
\bibliography{arxiv_eccv.bbl}
\newpage
\appendix
\section{\\More Examples} \label{App:AppendixA}

\begin{figure*}
\centering
\begin{tabular}{cc}
{\includegraphics[width=5.0cm]{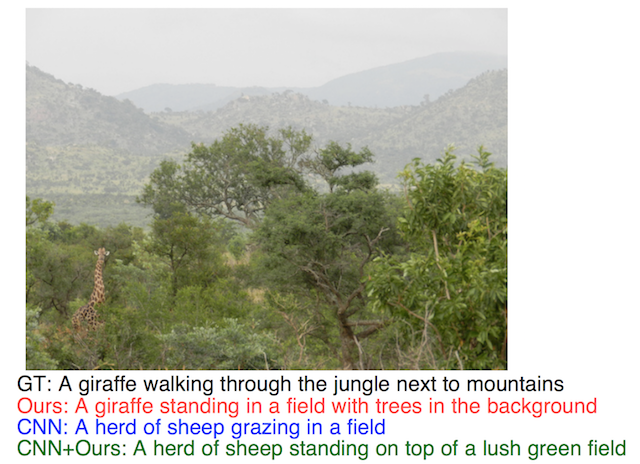}} \hfill& 
{\includegraphics[width=5.0cm]{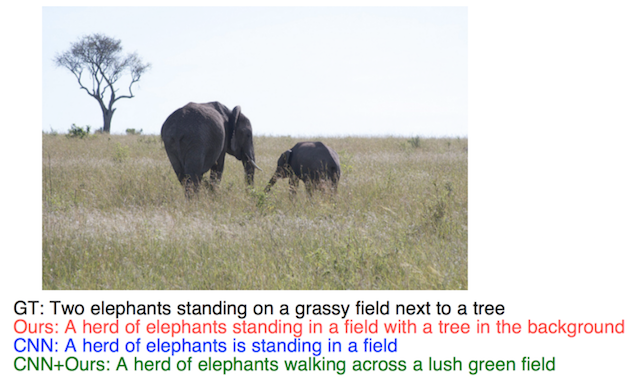}} \\ 
\end{tabular}
\label{fig:captions}
\vspace{-3ex}
\end{figure*}

\begin{figure*}
\centering
\begin{tabular}{cc}
{\includegraphics[width=4.2cm]{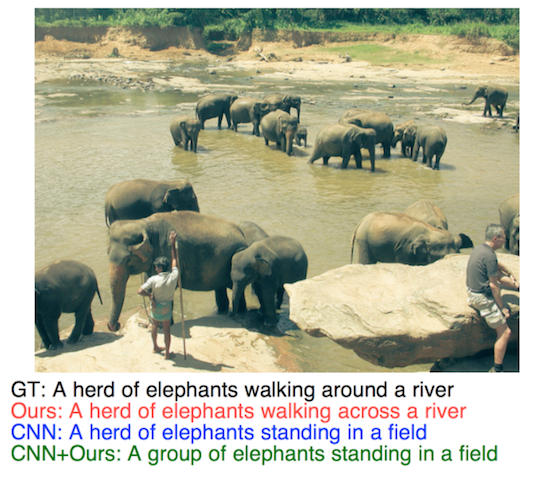}} \hfill&
{\includegraphics[width=4.5cm]{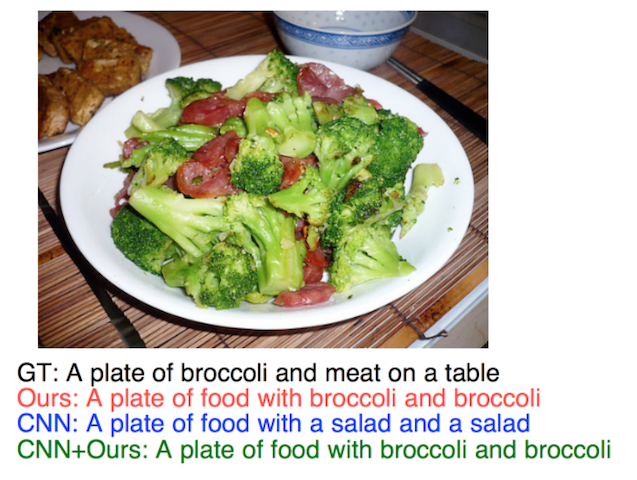}} \\
\end{tabular}
\label{fig:captions}
\vspace{-3ex}
\end{figure*}

\begin{figure*}
\centering
\begin{tabular}{cc}
{\includegraphics[width=5.0cm]{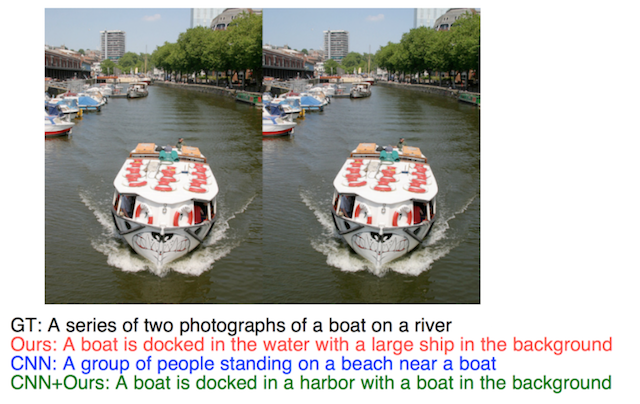}} \hfill&
{\includegraphics[width=5.0cm]{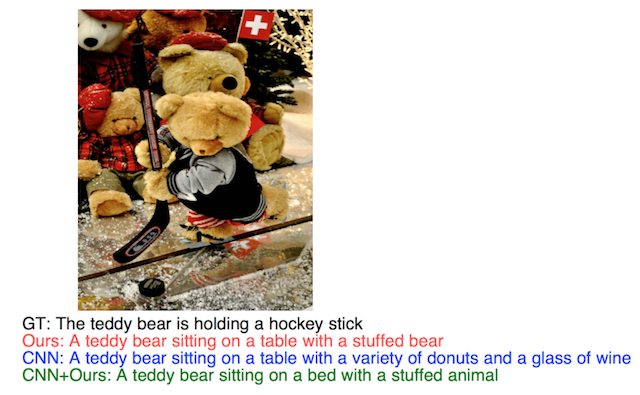}} \\
\end{tabular}
\label{fig:captions}
\vspace{-3ex}
\end{figure*}

\makeatletter
\setlength{\@fptop}{5pt}
\makeatother

\begin{figure*}[t!]
\centering
\begin{tabular}{cc}
{\includegraphics[width=5.0cm]{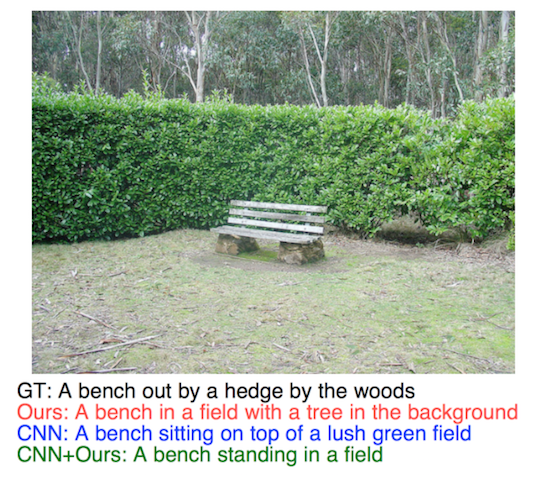}} \hspace{1.2cm}&
{\includegraphics[width=5.0cm]{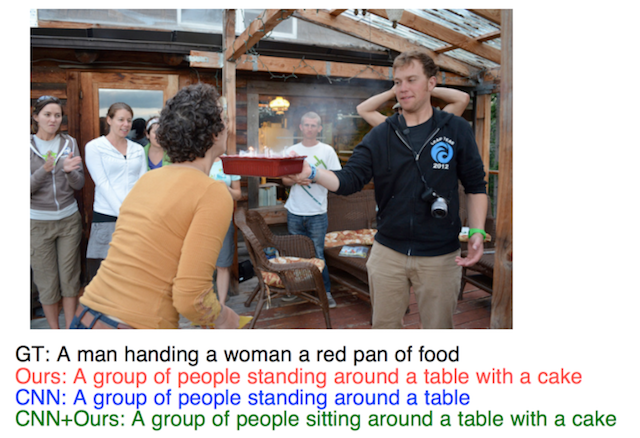}} \\
{\includegraphics[width=5.0cm]{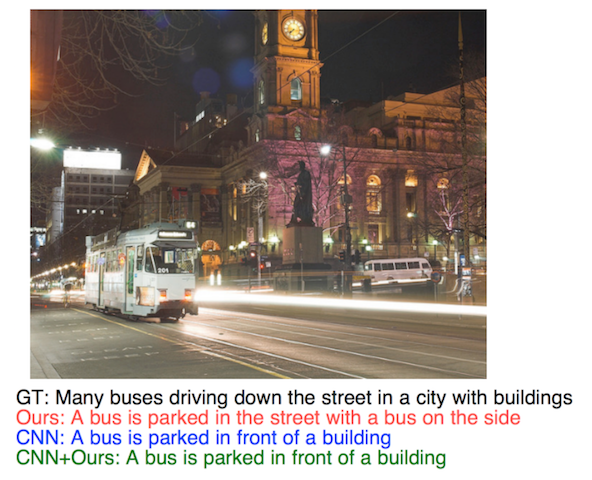}} \hspace{1.2cm}&
{\includegraphics[width=5.0cm]{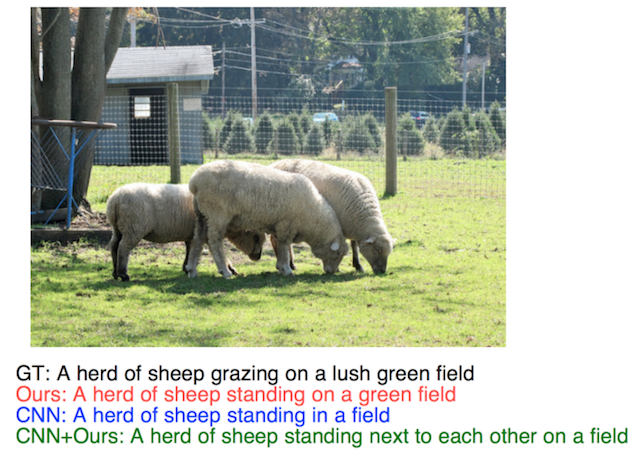}} \\
\end{tabular}
\label{fig:captions}
\vspace{-1ex}
\end{figure*}

\end{document}